\documentclass{article}
\usepackage[preprint]{spconf}
\usepackage{amsmath,graphicx}
\usepackage{wrapfig,lipsum,booktabs}
\usepackage{float}
\usepackage{hyperref}

\renewenvironment{thebibliography}[1]{
  \begin{oldthebibliography}{#1}
    \setlength{\itemsep}{0em}
    \setlength{\parskip}{0em}
}
{
  \end{oldthebibliography}
}

\copyrightnotice{978-1-5386-6249-6/19/\$31.00 \copyright 2019 IEEE}

\title{A Structurally Regularized Convolutional Neural Network for Image Classification using Wavelet-based SubBand Decomposition}
\name{Pavel~Sinha, Ioannis~Psaromiligkos, Zeljko~Zilic
\thanks{This research has benefited from a TechAccel grant, McGill University.}
\thanks{\copyright 2019 IEEE.  Personal use of this material is permitted.  Permission from IEEE must be obtained for all other uses, in any current or future media, including reprinting/republishing this material for advertising or promotional purposes, creating new collective works, for resale or redistribution to servers or lists, or reuse of any copyrighted component of this work in other works.}
\thanks{Published in the \emph{Proceedings of the 2019 IEEE International Conference on Image Processing (ICIP)}. DOI: \href{https://doi.org/10.1109/ICIP.2019.8804202}{10.1109/ICIP.2019.8804202}}
}
\address{McGill University, Department of Electrical \& Computer Engineering, Montreal, Canada\\
{pavel.sinha@mail.mcgill.ca, ioannis.psaromiligkos@mcgill.ca, zeljko.zilic@mcgill.ca}
}
\begin{document}
\maketitle
\begin{abstract}
We propose a convolutional neural network (CNN) architecture for image classification based on subband decomposition of the image using wavelets. The proposed architecture decomposes the input image spectra into multiple critically sampled subbands, extracts features using a single CNN per subband, and finally, performs classification by combining the extracted features using a fully connected layer. Processing each of the subbands by an individual CNN, thereby limiting the learning scope of each CNN to a single subband, imposes a form of structural regularization. This provides better generalization capability as seen by the presented results. The proposed architecture achieves best-in-class performance in terms of total multiply-add-accumulator operations and nearly best-in-class performance in terms of total parameters required, yet it maintains competitive classification performance. We also show the proposed architecture is more robust than the regular full-band CNN to noise caused by weight-and-bias quantization and input quantization.
\end{abstract}


\section{Introduction}
\label{sec:intro}

\par Deep learning has resulted in state-of-the-art performance in image recognition and vision tasks. 
Most of these achievements can be attributed to the use of convolutional neural networks (CNNs)~\cite{Lecun98gradient-basedlearning}. 
Since then, several other improvements to the CNN architecture have been proposed, including 
AlexNet~\cite{NIPS2012_4824}, 
VGG~\cite{DBLP:journals/corr/SimonyanZ14a}, 
GoogleNet~\cite{DBLP:journals/corr/SzegedyLJSRAEVR14}, 
ResNet~\cite{DBLP:journals/corr/HeZRS15}, 
Spatial Pyramid Pooling~\cite{DBLP:journals/pami/HeZR015}, 
SqueezeNet~\cite{DBLP:journals/corr/IandolaMAHDK16},
and more. 

\par The increasing complexity of CNNs poses challenges to state-of-the-art implementations. There are numerous techniques to reduce the computational cost of CNNs. 
Pruning of filters to simplify a CNN was proposed in~\cite{DBLP:journals/corr/LiKDSG16}. Another approach that used sparsity to reduce the number of filters per channel and per stage of a CNN was introduced in~\cite{DBLP:journals/corr/ChangpinyoSZ17}. SqueezeNet was introduced in~\cite{DBLP:journals/corr/IandolaMAHDK16} that claimed 50$\times$ fewer parameters than AlexNet, by using $1\times 1$ convolutional filters and reducing the overall number of parameters. Another work on model compression was introduced by~\cite{DBLP:journals/corr/SzeCYE17}. A characteristic shared by most of these methods is that they  can be reduced architecture-wise to special cases of the base CNN introduced in~\cite{NIPS2012_4824}. 

\par Deep networks require several layers of weights to be trained and even with millions of training data samples, overfitting remains inevitable~\cite{DBLP:journals/corr/PoggioMRML16}. Some recent techniques to combat overfitting include data augmentation~\cite{NIPS2012_4824}, weight regularization~\cite{NIPS2009_3848}, 
dropouts~\cite{NIPS2012_4824}, 
and adaptive regularization of weight vectors~\cite{NIPS2009_3848}. 
There is also a notion of \emph{structural regularization}, 
wherein constraints are imposed on the network structure rather than on the weight updates to limit overfitting~\cite{DBLP:journals/corr/Sun14c}.  Several works focus on this approach. A jointly enforced global wavelet domain sparsity constraint together with a learned analysis sparsity prior was introduced in~\cite{7362705}. 
A wavelet-regularized semi-supervised learning algorithm using suitably defined spline-like graph wavelets was introduced in~\cite{6736905}. In a recent work, a method of Graph-Spectral-Regularization was introduced in~\cite{2018arXiv181000424T}.

\par Multi-resolution analysis using wavelets was introduced by 
Daubechies~\cite{doi:10.1002/cpa.3160410705}.
It is well known that decomposing an image into subbands using wavelets is advantageous for image analysis. Not surprisingly, wavelets have been used with CNNs in several works. In  ~\cite{doi:10.1117/12.208730}, a single layer CNN was proposed in which the convolution kernel was wavelet-based. The model could not utilize the subbands from a regularization perspective and did not present an automated learning strategy as in deep learning. Another approach~\cite{DBLP:journals/corr/KanMY16} used wavelet decomposition for hierarchical image reconstruction to analyze CT scan images. 
A similar work on multi-resolution analysis with wavelets and CNNs was presented in~\cite{DBLP:journals/corr/abs-1805-08620}. The use of a scattering network as a generic and fixed initialization of the first layer in a CNN achieved similar results compared to learning the weights of the first layer was seen in~\cite{DBLP:journals/corr/OyallonBZ17}. 

In this paper, and in the context of image classification, we leverage subband decomposition to introduce a new structurally regularized CNN architecture wherein multiple CNNs are used to process the input image at different spatial scales as represented by the critically sampled and equally band-limited subbands obtained through a wavelet decomposition. The new architecture represents a departure from the ones used in the above-mentioned methods, where a single CNN was used to process the complete multi-scale wavelet output.
As we will see through qualitative arguments and extensive experimental studies the proposed architecture exhibits characteristics that translate to significant computational and classification performance advantages.

\begin{figure*}[t]
    \centering
    \includegraphics[scale=0.5]{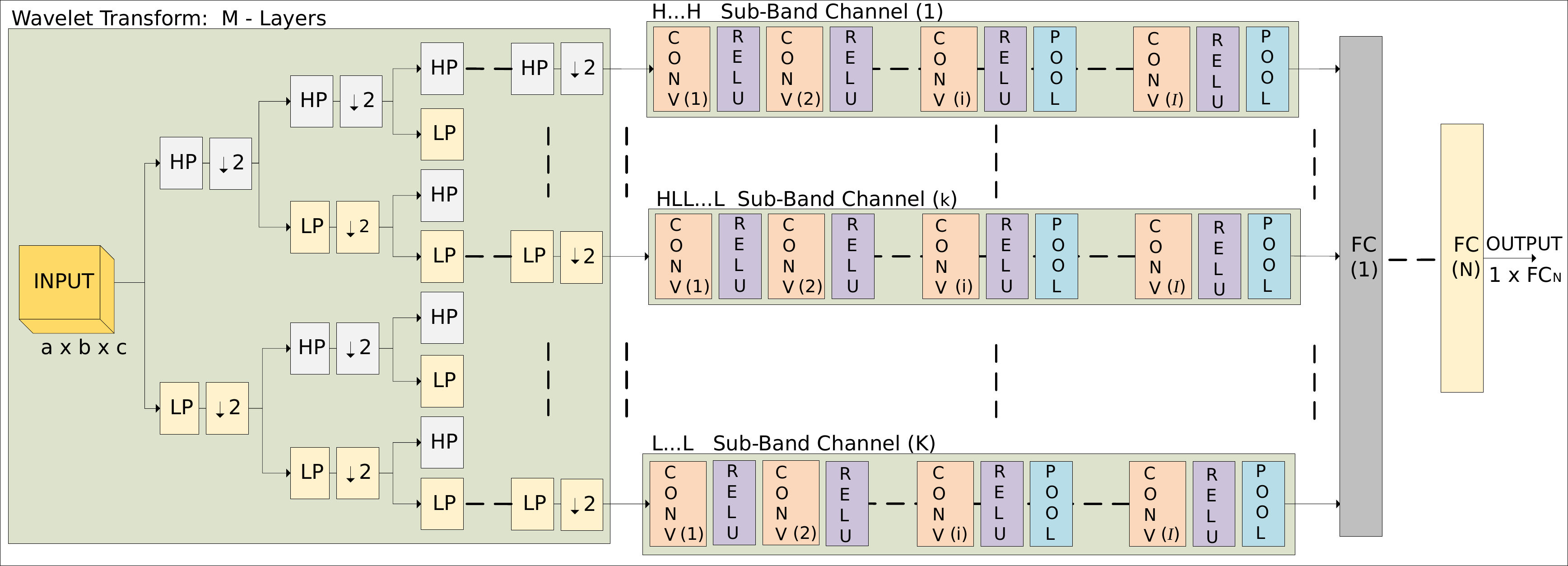}
    \vspace{-0.35cm}
    \caption{Architecture of an $M$-layer SRCNN, parametrized by input dimensions (a$\times$b$\times$c), number of subbands $K$, convolutional layers per subband $I$, FC layers $N$ and output classes ${\text{FC}}_{N}$, all open to optimization. The wavelet transform uses High-Pass (HP) and Low-Pass (LP) filters, followed by decimation of 2.}
    \label{fig_SRCNN}
    \vspace{-0.3cm} 
\end{figure*}

\section{Proposed Architecture}

\par The proposed architecture termed Subband Regularized CNN (SRCNN) is presented in Figure \ref{fig_SRCNN}. In the first stage, the input image is decomposed into subbands through a 2D discrete wavelet transform (2D-DWT). The  SRCNN architecture is based on processing each of the subbands separately by individual CNNs. The field of view of each CNN is hence restricted to a dedicated subband, making each CNN indifferent to the rest of the subbands. Importantly, this subband decomposition structure reduces the overall computational cost.  

 We represent the complete decomposition of the input image  $X_{\text{in}}$ into $K$ subbands by:
\vspace{-0.1cm}
\begin{flalign}
\,(X_0^{1}, \ldots, X_0^{K}) = \text{DWT} (X_{\text{in}}\,, K, M)
\label{eq_1}
\vspace{-0.4cm}
\end{flalign}
where $M$ is the number of DWT layers, $K$ is the number of subbands, and $X_0^{k}$ ($k=1, \ldots, K$) are the DWT coefficients for the $k^{th}$ subband. We have chosen the Daubechies (D2) family of basis functions for DWT~\cite{57199}.
This constitutes the simplest Daubechies wavelet basis, with a single vanishing moment. Being symmetric, they offer linear phase characteristics and 
do not suffer from edge effect characteristics of higher order wavelets~\cite{57199}.

\par The SRCNN architecture in Figure \ref{fig_SRCNN} is a generalized structure. The exact configuration of the  architecture implemented in this paper is given in Table \ref{table_arch_config}. 
The input image is first decomposed into $K$ subbands as described by Equation \ref{eq_1}. The subbands are then individually passed through their corresponding CNNs. Finally, the fully connected (FC) layers combine the feature outputs of the subband CNNs and perform image classification. The output of the CNN at the $k^{th}$ subband and $i^{th}$ layer is given by:
\begin{flalign}
\,{X_{i+1}^{k}} = \text{Pool}(\text{ReLU}(\text{Conv}(X_{i}^{k}\,, \,{W_{i}^{k}}) , \,{L_{i}}^{k}) , \,{P_{{i}}}^{k})
\label{eq_2}
\end{flalign}
where Conv represents the convolution between the input ${X_{i}}^{k}$ of the $i^{th}$ layer and the weights ${W_{i}}^{k}$. ReLU($\cdot$) indicates the ReLU activation function with $\,{L_{i}}^{k}$ representing the leakage percentage value~\cite{DBLP:journals/corr/XuWCL15} which is a real number between 0 and 1. Pool($\cdot$) represents the pooling function with pooling parameters ${P_{{i}}}^{k}$. The outputs of the subband CNNs are accumulated to yield ${X^{\text{FC}_0}}$ which is the input to the first FC layer:
\begin{flalign}
\,X^{\text{FC}_0} = ({X_I}^{1}, \, . \,. \,. \,,{X_I}^{K})
\label{eq_3}
\end{flalign}
where $I$ is the number of layers in the subband CNNs. The output at each FC layer is given by:
\begin{flalign}
\,X^{\text{FC}_{n+1}} = \text{ReLU}(\,W^{\text{FC}_n} \,\cdot\, X^{\text{FC}_n} \,,\, \,L^{\text{FC}_n})
\label{eq_4}
\end{flalign}
where $X^{\text{FC}_{n}}$ denotes the output of the $n^{th}$ FC layer, $\cdot$ indicates matrix multiplication and $L^{\text{FC}_n} $ indicates ReLU leakage value. Finally, the output of the last FC layer $X^{{\text{FC}}_{N}}$, indexed by $N$, produces the 
SRCNN's output $Y$. Equations \ref{eq_1} to \ref{eq_4} describe the complete input to output relation of the proposed subband based CNN.

\section{Properties of the Architecture}

\par The proposed architecture emphasizes regularization through its structure, thus it is \emph{structurally regularized}. To enhance regularization effectiveness, the decomposed subbands are critically sampled and band-limited before being processed by individual subband CNNs. Each of the subband CNNs is inhibited from accessing information across the entire spectrum of the input.  Overall, each of the CNNs cannot learn sample-specific features present in the entire spectrum of the input. This restriction combined with weight regularization within each CNN, improves regularization, leading to better generalization ability and reduced overfitting, as demonstrated by the accuracy performance comparison in Table \ref{table_compute_benefit}. Apart from accuracy, the difference in Top-5 and Top-1 accuracy result can be considered as an indicator for generalization effectiveness. A lower difference value indicates better generalization which, in our case, outperforms other state-of-the-art networks.

\par The lossless decomposition of the input spectrum into orthogonal subbands allows isolated analysis of the spatial representation of each subband.  This is beneficial in the case of corrupted images. Indeed, corruption of the input image by noise, deformities from lens aberration, incorrect exposure, low lighting, etc., does not affect the entire spectrum equally; in reality, some subbands are corrupted more. Isolating the subbands ensures that the corruption of extracted features is limited to the affected subbands, as opposed to a full-band CNN that considers the entire spectrum for feature extraction. 

\par Along similar lines, quantization noise in each weight is confined within the subband and does not affect the entire spectrum. In contrast, in a regular CNN, quantization noise in any weight can potentially corrupt the entire spectrum, since quantization noise can have large bandwidth. Results indicate that compared to full-band CNN, SRCNN proves more robust to weight and input quantization.

\par The subband decomposition also introduces high degree of sparsity in the subbands, specifically in the non-basebands containing mostly edge information. This sparsity is introduced at the very input of the subband CNNs. It is well known that sparse inputs help reduce CNN complexity \cite{DBLP:journals/corr/ChangpinyoSZ17}. 

\par Random initialization of weights when training a full-band CNN does not guarantee scanning of the entire spectrum for useful features. In the proposed structure, the CNNs focus only on their corresponding subbands hence the entire spectral decomposed into subbands is covered equally. 

\par The decomposition reduces the input spatial dimension along rows and columns by $2^M$ each, where $M$ is the number of decomposition layers. The total reduction of input dimension is effectively on the order of $4^M$ for two-dimensional input data such as images. The convolution operation accounts for the bulk of computations in a CNN. The total computation cost depends super-linearly on the size of the convolution filters~\cite{DBLP:conf/cvpr/He015} and the sample point counts per dimension, all of which are significantly reduced in our case.

\par The subband decomposition architecture offers parallel computation along each subband. The parallelism also provides a mechanism to reduce internal memory footprint by sequentially computing each subband and reusing internal scratch memory to compute each subband CNN.

\par Finally, decomposition of input spectra into subbands is a generalized technique and can be applied to any CNN to improve regularization and thereby improve generalization capacity and improve overall performance.

\section{Experimental Setup and Results}
\subsection{Methodology and Training}
\par We use MNIST, CIFAR-10/100, Caltech-101 and ImageNet-2012 datasets (used by~\cite{NIPS2012_4824}) in our experimentation.
We compare the proposed architecture against two benchmarks: (i) a full-spectrum base CNN (BCNN) model that closely resembles AlexNet~\cite{NIPS2012_4824} and VGG-16~\cite{DBLP:journals/corr/SimonyanZ14a}; 
(ii) the Transform CNN (TCNN) architecture which shares the same wavelet front-end as SRCNN, except that the subbands are combined and processed by a single CNN with the same number of weights and layers as BCNN. A single layer, subband decomposed TCNN (4 subbands) with 3 input color channels will result in a total input of 12 channels and with $1/2$ the length and width of the original input image. Table~\ref{table_arch_config} shows the parameters of the models used. To study the effect of learning in subband domain, we keep most of the parameters constant across BCNN, SRCNN and TCNN. 
We compare the number of MAC operations, total number of parameters and accuracy with several well known state-of-the-art CNN architectures. 

\indent  We train using stochastic gradient descent (SGD) with a mini batch size of 64, batch normalized, randomly picked images per mini batch, momentum of 0.9 and weight decay of 0.0005~\cite{NIPS2012_4824}. The update equations for $W_i^{k}$ 
are:
\begin{flalign}
W_i^{k}(l+1) &= W_i^{k}(l) + V_i^{k}(l+1)\\ 
V_i^{k}(l+1) &= 0.9\,V_i^{k}(l)-0.0005\,\epsilon\,W_i^{k}(l) -\, \epsilon\,  \left.\overline{\frac{\partial L}{\partial w}} \right| _{W_i^{k}(l)}
\label{eq_6}
\end{flalign}
Here, $l$ is the iteration index, $V_i^{k}(l)$ is the momentum at the $l^{th}$ iteration and $k^{th}$ subband, $\epsilon$ the learning rate, and $\left.\overline{\frac{\partial L}{\partial W}} \right| _{W_i^{k}(l)} $ is the average over the $l^{th}$ batch of the derivative of the objective function with respect to $W_i^k$, evaluated at $W_i^{k}(l)$. We initialize the learning rate to 0.01 and all biases to 1, while 
we initialize the weights by drawing from a Gaussian distribution with a standard deviation of 0.01. 

\begin{table}[t]
\caption{
CNN architectural configuration used for BCNN, TCNN and SRCNN. 
Every convolutional layer is followed by a leaky ReLU~\protect\cite{DBLP:journals/corr/XuWCL15} with 10$\%$ leakage value. 
}
%\vspace{0.2cm}
\centering
\label{table_arch_config}
\scalebox{0.61}{
\begin{tabular}{|l|l|l|}
\hline
\textbf{Dataset} & MNIST Or CIFAR-10/100 & \begin{tabular}[c]{@{}l@{}} Caltech-101 Or ImageNet-2012\end{tabular} \\ \hline
\textbf{Architectures} & BCNN / TCNN / SRCNN & BCNN / TCNN  / SRCNN \\ \hline
\textbf{Input Size} & 28x28x1 Or 32x32x3 & \begin{tabular}[c]{@{}l@{}}224x224x3 \end{tabular} \\ \hline
\textbf{SubBand} & - / 1-Layer / 1-Layer & -  /  1-Layer  /  1-Layer \\ \hline
\textbf{CONV+ReLu} & \begin{tabular}[c]{@{}l@{}}3x3x1x64 / 3x3x4x64 / \\ 3x3x1x16x4\end{tabular} & \begin{tabular}[c]{@{}l@{}}3x3x3x64 / 3x3x12x64  /   3x3x3x16x4\end{tabular} \\ \hline
\textbf{CONV+ReLu} & \begin{tabular}[c]{@{}l@{}}3x3x64x128 / 3x3x64x128 /\\ 3x3x16x32x4\end{tabular} & \begin{tabular}[c]{@{}l@{}}3x3x64x64  /   3x3x64x64  /   3x3x16x16x4\end{tabular} \\ \hline
\textbf{CONV+ReLu} & \begin{tabular}[c]{@{}l@{}}3x3x128x256 /  3x3x128x256 /\\ 3x3x32x64x4\end{tabular} & \begin{tabular}[c]{@{}l@{}}3x3x64x64  /   3x3x64x64  /   3x3x16x16x4\end{tabular} \\ \hline
\textbf{CONV+ReLu} & - & \begin{tabular}[c]{@{}l@{}}3x3x64x64  /   3x3x64x64  /   3x3x16x16x4\end{tabular} \\ \hline
\textbf{CONV+ReLu} & - & \begin{tabular}[c]{@{}l@{}}3x3x64x64  /   3x3x64x64  /   3x3x16x16x4\end{tabular} \\ \hline
\textbf{POOL} & 2-by-2 & 2-by-2 \\ \hline
\textbf{CONV+ReLu} & \begin{tabular}[c]{@{}l@{}}3x3x256x512 /  3x3x256x512 /\\ 3x3x64x128x4\end{tabular} & \begin{tabular}[c]{@{}l@{}}3x3x64x128  /  3x3x64x128 /  3x3x16x32x4\end{tabular} \\ \hline
\textbf{CONV+ReLu} & \begin{tabular}[c]{@{}l@{}}3x3x512x128 / 3x3x512x128 / \\ 3x3x128x32x4\end{tabular} & \begin{tabular}[c]{@{}l@{}}3x3x64x128  /  3x3x64x128  /  3x3x16x32x4\end{tabular} \\ \hline
\textbf{CONV+ReLu} & - & \begin{tabular}[c]{@{}l@{}}3x3x64x128 /  3x3x64x128 /  3x3x16x128x4\end{tabular} \\ \hline
\textbf{CONV+ReLu} & - & \begin{tabular}[c]{@{}l@{}}3x3x64x128 /  3x3x64x128 /  3x3x16x32x4\end{tabular} \\ \hline
\textbf{CONV+ReLu} & - & \begin{tabular}[c]{@{}l@{}}3x3x64x128 /  3x3x64x128 /  3x3x16x32x4\end{tabular} \\ \hline
\textbf{POOL} & 2-by-2  & 2-by-2 \\ \hline
\textbf{CONV+ReLu} & - & \begin{tabular}[c]{@{}l@{}}3x3x64x128 /  3x3x64x128 /  3x3x16x32x4\end{tabular} \\ \hline
\textbf{CONV+ReLu} & - & 3x3x128x128 \\ \hline
\textbf{CONV+ReLu} & - & 3x3x128x128 \\ \hline
\textbf{CONV+ReLu} & - & 3x3x128x128 \\ \hline
\textbf{CONV+ReLu} & - & 3x3x128x128 \\ \hline
\textbf{POOL} & - & 2-by-2 \\ \hline
\textbf{FC-1} & 4x4x128x4096 & \begin{tabular}[c]{@{}l@{}}4x4x128x4096  \end{tabular} \\ \hline
\textbf{DROPOUT}~\cite{NIPS2012_4824} & 50\% & 50\% \\ \hline
\textbf{FC-2} & 4096x1024 & \begin{tabular}[c]{@{}l@{}}4096x1024(C.Tech) Or 4096x4096(Im.Net)\end{tabular} \\ \hline
\textbf{DROPOUT}~\cite{NIPS2012_4824} & 50\% & 50\% \\ \hline
\textbf{FC-3} & 1024x10 / 1024x100 & \begin{tabular}[c]{@{}l@{}}4096x102 / 4096x1000 \end{tabular} \\ \hline
\textbf{SOFTMAX} & 1x10 / 1x100 & \begin{tabular}[c]{@{}l@{}}1x102 / 1x1000\end{tabular} \\ \hline
\end{tabular}}
\vspace{-0.5cm}
\end{table}

\begin{table}[t]
\centering
\caption{Comparison of total MAC operations, parameters used and classification accuracy of 1\&2-layer DWT SRCNN architecture with other well established CNN models for the ImageNet-2012 dataset.}
\label{table_compute_benefit}
\scalebox{0.648}{
\begin{tabular}{|c|c|c|c|c|c|c|}
\hline
\textbf{Models} & \textbf{MACs} & \textbf{\begin{tabular}[c]{@{}c@{}}Param.\\ (Million)\end{tabular}} & \textbf{\begin{tabular}[c]{@{}c@{}}Param.\\ (MByte)\end{tabular}} & \textbf{\begin{tabular}[c]{@{}c@{}}Accuracy\\ (Top-1)\end{tabular}} & \textbf{\begin{tabular}[c]{@{}c@{}}Accuracy\\ (Top-5)\end{tabular}} & \textbf{\begin{tabular}[c]{@{}c@{}}Delta \\ Top (5 - 1)\end{tabular}} \\ \hline
MobileNet V1 & 569 M & 4.24 & 2 & 70.9 & 89.9 & 19 \\ \hline
MobileNet V2 & 300 M & 3.47 & 1.7 & 71.8 & 91 & 19.2 \\ \hline
Google Net & 741 M & 6.99 & 3.3 & - & 92.1 & - \\ \hline
AlexNet & 724 M & 60.95 & 29.1 & 62.5 & 83 & 20.5 \\ \hline
SqueezeNet & 451 M & \textbf{1.24} & \textbf{0.6} & 57.5 & 80.3 & 22.8 \\ \hline
ResNet-50 & 3.9 B & 25.6 & 12.2 & \textbf{75.2} & \textbf{93} & 17.8 \\ \hline
VGG & 15.5 B & 138 & 65.8 & 70.5 & 91.2 & 20.7 \\ \hline
Inception-V1 & 1.43 B & 7 & 3.3 & 69.8 & 89.3 & 19.5 \\ \hline
SRCNN (1L) & \textbf{169.5 M} & 42.05 & 20.1 & 65.6 & 82.17 & \textbf{16.57} \\ \hline
SRCNN (2L) & \textbf{46.34 M} & 13.64 & 6.5 & - & - & - \\ \hline
\end{tabular}
}
\vspace{-0.568cm}
\end{table}

\begin{table}[t]
\caption{Classification accuracy with 1-layer DWT architecture with parameters indicated in Table~\ref{table_arch_config}.}
\centering
\label{acc_percent}
\scalebox{0.63}{
\begin{tabular}{|c|c|c|c|c|}
\hline
\textbf{Dataset}                                               & \textbf{BCNN}  & \textbf{TCNN}  & \textbf{SRCNN}          & \begin{tabular}[c]{@{}c@{}}\textbf{State-of-art}\end{tabular}      \\ \hline
MNIST                                                 & 99.72 & 99.76 & \textbf{99.83} & \textit{99.79}~\cite{pmlr-v28-wan13} \\ \hline
CIFAR-10                                              & 95.37 & 96.59 & \textbf{96.71} & \textit{96.53}~\cite{DBLP:journals/corr/Graham14a} \\ \hline
CIFAR-100                                             & 80.72 & 81.74 & \textbf{82.97} & \textit{81.70 \cite{DBLP:journals/corr/ZagoruykoK16} } \\ \hline
\begin{tabular}[c]{@{}c@{}}CALTECH-101\end{tabular} & 82.17  & 83.89  & 86.93      & \begin{tabular}[c]{@{}c@{}}\textbf{89.47}(ZF-5)~\cite{DBLP:journals/pami/HeZR015}\end{tabular} \\ \hline
\end{tabular}}
\vspace{-0.3cm}
\end{table}

\begin{table}[t]
\vspace{-0.3cm}
\caption{Classification accuracy with input quantization for 1-layer DWT architecture.}
\centering
\label{input_quant}
\scalebox{0.63}{
\begin{tabular}{|c|c|c|c|c|c|c|}
\hline
\textbf{Datasets}                                                               & \textbf{Models} & \textbf{1-bit} & \textbf{2-bits} & \textbf{4-bits} & \textbf{6-bits} & \textbf{8-bits}         \\ \hline
{MNIST}                                                 & BCNN   & 97.42 & 97.53  & 97.6   & 97.69  & 99.72          \\ \cline{2-7} 
                                                                       & TCNN   & 97.87 & 98.03  & 99.45  & 99.53  & 99.76          \\ \cline{2-7} 
                                                                       & SRCNN  & \textbf{99.5}  & \textbf{99.6}   & \textbf{99.63}  & \textbf{99.64}  & \textbf{99.83} \\ \hline
{\begin{tabular}[c]{@{}c@{}}CIFAR-10\end{tabular}}    
& BCNN   & 76.36 & 76.06  & 76.39  & 85.66  & 95.37          \\ \cline{2-7} 
& TCNN   & 76.66 & 76.38  & 79.2   & 89.56  & 96.59          \\ \cline{2-7} 
& SRCNN  & \textbf{78.13} & \textbf{79.2}   & \textbf{80.03}  & \textbf{91.32}  & \textbf{96.71} \\ \hline
{\begin{tabular}[c]{@{}c@{}}CIFAR-100\end{tabular}}   
& BCNN   & 54.89 & 59.66  & 68.8   & 73.85  & 80.72          \\ \cline{2-7} 
& TCNN   & 58.91 & 61.03  & \textbf{69.87}  & \textbf{74.75}  & 81.74 \\ \cline{2-7} 
& SRCNN  & \textbf{60.14} & \textbf{64.79}  & 69.15  & 74.07  & \textbf{82.97} \\ \hline
{\begin{tabular}[c]{@{}c@{}}CALTECH-101\end{tabular}} & BCNN   & 69.87 & 71.96 & 76.2    & 79.31  & 82.17          \\ \cline{2-7} 
                                                                       & TCNN   & \textbf{71.66}  & \textbf{76.29}  & \textbf{80.91}  & 81.47  & 83.89          \\ \cline{2-7} 
                                                                       & SRCNN  & 71.17 & 74.94  & 77.61  & \textbf{83.16}  & \textbf{86.93}          \\ \hline
\end{tabular}}
\vspace{-0.5cm}
\end{table}

\begin{table}[!]
\caption{Classification accuracy with weight-and-bias quantization for 1-Layer DWT architecture.}
\centering
\label{weight_quant}
\scalebox{0.63}{
\begin{tabular}{|l|l|l|l|l|l|l|}
\hline
\textbf{Datasets} & \multicolumn{3}{l|}{MNIST}     & \multicolumn{3}{l|}{CIFAR-10}    \\ \hline
\textbf{Models}   & BCNN  & TCNN  & SRCNN          & BCNN   & TCNN   & SRCNN          \\ \hline
\textbf{8-bits}   & 90.3  & 90.91 & \textbf{91.65} & 60.85  & 61.37  & \textbf{63.54} \\ \hline
\textbf{16-bits}  & 97.53 & 97.9  & \textbf{99.65} & 76.13  & 79.46  & \textbf{79.84} \\ \hline
\textbf{32-bits}  & 99.72 & 99.76 & \textbf{99.83} & 95.37  & 96.59  & \textbf{96.71} \\ \hline
\textbf{Datasets} & \multicolumn{3}{l|}{CIFAR-100} & \multicolumn{3}{l|}{Caltech-101} \\ \hline
\textbf{Models}   & BCNN  & TCNN  & SRCNN          & BCNN   & TCNN   & SRCNN          \\ \hline
\textbf{8-bits}   & 56.17 & 61.13 & \textbf{63.15} & 71.13  & 75.4   & \textbf{82.35} \\ \hline
\textbf{16-bits}  & 53.78 & 61.13 & \textbf{61.37} & 79.67  & 80.01  & \textbf{83.16} \\ \hline
\textbf{32-bits}  & 80.72 & 81.74 & \textbf{82.97} & 82.17  & 83.89  & \textbf{86.93} \\ \hline
\end{tabular}}
\vspace{-0.5cm}
\end{table}

\subsection{Results}

\noindent\textbf{Classification Accuracy:}  Table \ref{acc_percent} summarizes the classification accuracy results.
As we can see, the 1-Layer SRCNN improves the state-of-the-art performance for MNIST, CIFAR-10 and CIFAR-100 datasets by a fair margin. Replacing the 1-layer DWT with a 2-layer DWT decomposition, i.e., with a 2-layer subband decomposition or 16 subbands, we achieve an accuracy of 84.37\% for TCNN and 88.93\% for SRCNN, on the Caltech-101 dataset. With a 1-layer subband decomposition, SRCNN trained on ImageNet-2012, we achieve top-5 and top-1 validation set accuracy~\cite{NIPS2012_4824} of 82.17\% and 65.6\%, respectively. Table \ref{table_compute_benefit} indicates that our proposed architecture achieves accuracy which is competitive with other state-of-the-art CNN networks that are heavily optimized.

\noindent\textbf{Computational Cost:}
Table \ref{table_compute_benefit} compares the total number of multiply-and-accumulate (MAC) operations and parameters used by state-of-the-art CNNs. Both the 1-layer and 2-layer subband decomposed SRCNNs perform best in class in terms of number of total MAC operations needed. On the number of parameters front, the SRCNN architecture performs fairly. However, the parameters of all compared CNNs fall between 0.6 to 65.8 MBytes, with the 1-layer and 2-layer SRCNN architectures at 20.1 and 6.5 MBytes, respectively. In practice, the difference between 6.5 and 0.6 MBytes can be ignored, where as a 10$\times$ reduction in total MAC operations can significantly improve computation time.

\noindent\textbf{Quantization Effects:}
The effect of input-data quantization and weights-and-biases quantization on classification accuracy is listed in Table \ref{input_quant} and Table \ref{weight_quant}, respectively, for BCNN, TCNN and SRCNN architectures. MNIST, CIFAR-10/100 and Caltech-101 datasets are native 8 bits per color. We quantize the input image to 1, 2, 4, 6 and 8 bits per color, and the weights and biases to 8, 16 and 32 bits IEEE floating point values. As we can see the SRCNN architecture is more robust than BCNN and TCNN.

\section{Conclusion}

\par The proposed SRCNN architecture achieves state-of-the-art performance with computation cost less than 10\% of an equivalent CNN. Our method owes its performance to structural regularization -- the input signal is losslessly decomposed into subbands and the subband CNNs are restrained from learning features in the other subbands, thereby reducing the risk of overfitting. In addition to computational benefits, the distribution of information across different subbands may vary greatly from class to class. As a result, the FC layers of SRCNN have more information compared to FC layers of a full-band CNN to separate the output space.  Further, noise and deformities are isolated to each subband and do not corrupt the rest, making classification robust compared to analyzing the entire spatial representation by a single CNN. Our architecture is also robust to input data quantization and weight-bias quantization error, which is critical in real life CNN applications where quantization is inevitable.

\bibliographystyle{IEEEbib}
\bibliography{my_bib}

\end{document}